\documentclass[runningheads]{llncs}

\usepackage{eccv}

\usepackage{eccvabbrv}
\usepackage{colortbl}
\usepackage{array}
\usepackage{multirow}
\usepackage{comment}
\usepackage{rotating}
\usepackage{adjustbox}
\usepackage{booktabs} 
\usepackage{siunitx}  
\usepackage{graphicx}
\usepackage{booktabs}
\definecolor{Blue74}{RGB}{8, 48, 107}
\definecolor{Blue41}{RGB}{178, 210, 231}
\definecolor{Blue26}{RGB}{247, 250, 255}
\definecolor{Blue59}{RGB}{49, 129, 190}
\definecolor{Blue82}{RGB}{8, 48, 107}
\definecolor{Blue66}{RGB}{32, 112, 180}
\definecolor{Blue18}{RGB}{247, 251, 255}
\definecolor{Blue34}{RGB}{198, 219, 239}
\definecolor{Blue71}{RGB}{8, 48, 107}
\definecolor{Blue36}{RGB}{214, 230, 244}
\definecolor{Blue29}{RGB}{247, 251, 255}
\definecolor{Blue64}{RGB}{16, 91, 164}
\definecolor{Blue83}{RGB}{8, 48, 107}
\definecolor{Blue70}{RGB}{22, 99, 170}
\definecolor{Blue17}{RGB}{247, 251, 255}
\definecolor{Blue30}{RGB}{208, 226, 242}
\usepackage[accsupp]{axessibility}  

\usepackage[pagebackref,breaklinks,colorlinks,citecolor=eccvblue]{hyperref}

\usepackage{orcidlink}

\begin{document}

\title{GenerateCT: Text-Conditional Generation of 3D Chest CT Volumes} 

\titlerunning{GenerateCT}

\author{\small Ibrahim Ethem Hamamci\inst{1} \and
Sezgin Er\inst{2} \and
Anjany Sekuboyina\inst{1} \and
Enis Simsar\inst{3} \and
Alperen Tezcan\inst{2} \and
Ayse Gulnihan Simsek\inst{2} \and
Sevval Nil Esirgun\inst{2} \and
Furkan Almas\inst{2} \and
Irem Doğan\inst{2} \and
Muhammed Furkan Dasdelen\inst{2} \and
Chinmay Prabhakar\inst{1} \and
Hadrien Reynaud\inst{4} \and
Sarthak Pati\inst{5} \and
Christian Bluethgen\inst{6} \and
Mehmet Kemal Ozdemir\inst{2} \and
Bjoern Menze\inst{1}}

\authorrunning{Hamamci et al.}

\institute{\begin{tabular}{c}
{$^{1}$University of Zurich \quad $^{2}$Istanbul Medipol University \quad $^{3}$ETH Zurich} \\
{$^{4}$Imperial College London \quad $^{5}$University of Pennsylvania \quad $^{6}$Stanford University}
\end{tabular}\\
\email{ibrahim.hamamci@uzh.ch}}

\maketitle

\begin{abstract}
Text-conditional medical image generation is vital for radiology, augmenting small datasets, preserving data privacy, and enabling patient-specific modeling. However, its applications in 3D medical imaging, such as CT and MRI, which are crucial for critical care, remain unexplored. In this paper, we introduce \textbf{GenerateCT, the first approach to generating 3D medical imaging conditioned on free-form medical text prompts}. GenerateCT incorporates a text encoder and three key components: a novel causal vision transformer for encoding 3D CT volumes, a text-image transformer for aligning CT and text tokens, and a text-conditional super-resolution diffusion model. Without directly comparable methods in 3D medical imaging, we benchmarked GenerateCT against cutting-edge methods, demonstrating its superiority across all key metrics. Importantly, we explored GenerateCT's clinical applications by evaluating its utility in a multi-abnormality classification task. First, we established a baseline by training a multi-abnormality classifier on our real dataset. To further assess the model's generalization to external datasets and its performance with unseen prompts in a zero-shot scenario, we employed an external dataset to train the classifier, setting an additional benchmark. We conducted two experiments in which we doubled the training datasets by synthesizing an equal number of volumes for each set using GenerateCT. The first experiment demonstrated an $11\%$ improvement in the AP score when training the classifier jointly on real and generated volumes. The second experiment showed a $7\%$ improvement when training on both real and generated volumes based on unseen prompts. Moreover, GenerateCT enables the scaling of synthetic training datasets to arbitrary sizes. As an example, we generated 100,000 3D CT volumes, fivefold the number in our real dataset, and trained the classifier exclusively on these synthetic volumes. Impressively, this classifier surpassed the performance of the one trained on all available real data by a margin of $8\%$. Lastly, domain experts evaluated the generated volumes, confirming a high degree of alignment with the text prompts. Access our code, model weights, training data, and generated data at \url{https://github.com/ibrahimethemhamamci/GenerateCT}.
\end{abstract}

\begin{figure}
    \centering
    \includegraphics[width=1\textwidth]{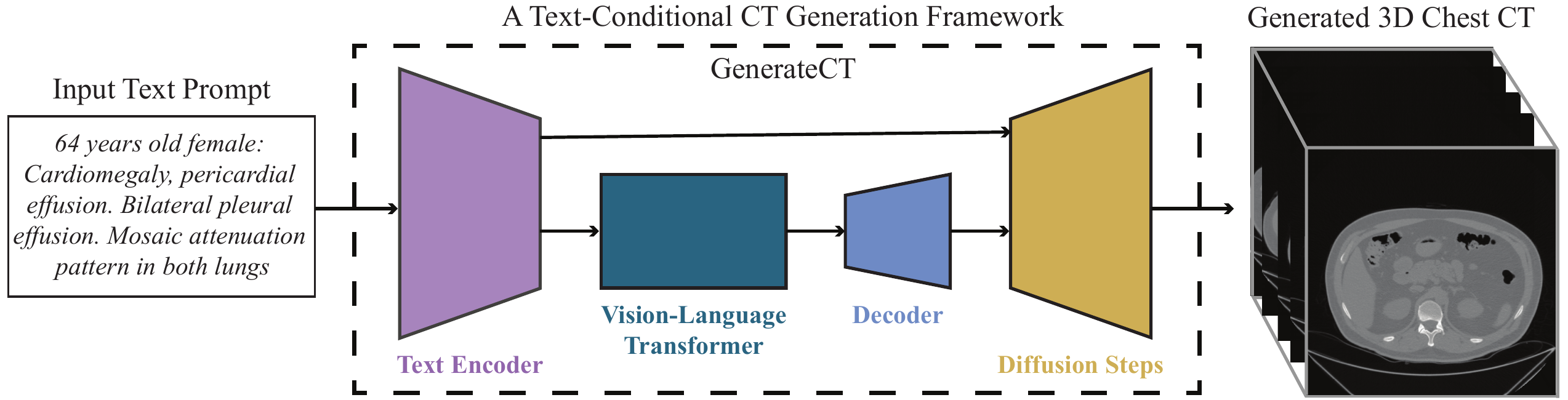}
    \caption{GenerateCT is a cascaded framework that generates high-resolution and high-fidelity 3D chest CT volumes based on medical language text prompts.}
    \label{first}
\end{figure}

\section{Introduction}
\label{sec:intro}

The text-conditional generation of synthetic images holds significant promise for the medical field by producing text-aligned and clinically-ready images, bypassing the need for labeling. It facilitates the large-scale adoption of machine learning, enhancing radiological workflows, accelerating medical research, and improving patient care. Additionally, it addresses key challenges in medical image analysis, such as data scarcity, patient privacy concerns, imbalanced class distribution, and the need for trained clinicians for manual annotation~\cite{lee2023unified}.

The field of natural image generation from free-flowing text has seen remarkable progress~\cite{ding2021cogview, ramesh2022hierarchical, rombach2022high, saharia2022photorealistic, yu2022scaling, balaji2022ediffi, chen2022re, gu2022vector, ho2022cascaded, nichol2021glide}. Despite these advancements, the medical field has yet to fully capitalize on the potential of generative modeling, due to the significant distribution shift between natural and medical images—and even among different medical domains~\cite{kebaili2023deep}. A particular application is the generation of 2D chest X-rays from radiology reports~\cite{chambon2022roentgen}, achieved by fine-tuning a pre-trained, open-source text-to-image model~\cite{rombach2022high}. However, extending this method to more spatially complex modalities, such as 3D computed tomography (CT) or magnetic resonance imaging (MRI), remains unexplored. The primary challenge is the exponential increase in computational complexity associated with the nature of 3D medical imaging~\cite{clark2019adversarial}. Furthermore, unlike in 2D generation, there are no pre-trained 3D models available for fine-tuning~\cite{zhang2023text}. Generating 2D slices instead of 3D volumes also poses significant challenges in the medical field due to the lack of spatial context. Additionally, the scarcity of 3D medical imaging data paired with radiology reports limits development~\cite{linna2022applications}.

Recognizing this gap, we propose \textbf{GenerateCT}, the first method for the synthesis of 3D medical imaging conditioned on free-form text prompts~(see~\Cref{first}), specifically targeting high-resolution 3D chest CT volumes. Our framework consists of three key components: The first is a novel causal vision transformer, CT-ViT, which encodes the 3D CT volumes into tokens. CT-ViT is trained to reconstruct 3D CT volumes autoregressively, allowing us to maintain high axial resolution and generate a variable number of axial slices, thus providing a variable axial field-of-view \cite{arnab2021vivit}. Second, a bidirectional text-image transformer aligns the CT tokens with the encoded tokens of the free-form radiology text. This alignment is facilitated using masked CT token prediction~\cite{chang2022maskgit}. Third, we employ a cascaded diffusion model~\cite{ho2022cascaded} to enhance the in-plane resolution of the generated low-resolution volumes. This step is also text-conditioned, ensuring faithful resolution enhancement based on the input prompt~\cite{saharia2022photorealistic}.

GenerateCT's uniqueness, being the first of its kind in 3D medical imaging, means that no directly comparable methods exist, further highlighting its novelty. Regardless, to demonstrate the effectiveness of our framework, we have designed some baseline methods using state-of-the-art generation models. First, to show the importance of 3D generation architecture for ensuring consistency in 3D chest CT volumes, we employ two text-conditional 2D image generation methods for comparison. We also implement a text-to-video generation model for 3D chest CT synthesis to highlight our framework's optimized benefits over other 3D generation approaches. Furthermore, we perform a comprehensive ablation study to underscore the effectiveness of GenerateCT's cascaded architecture.

GenerateCT synthesizes text-aligned 3D chest CT volumes, bypassing the need for labeling. Since GenerateCT is the first of its kind, we, more importantly, assessed its clinical utility in a multi-abnormality classification task. Initially, we established a baseline training classifier on all available real volumes. We expanded our training data by creating an equivalent number of synthetic volumes with GenerateCT, yielding an 11\% improvement in mean AP by training on this joint dataset. Furthermore, GenerateCT allowed us to massively scale our synthetic dataset; we produced 100,000 3D CT volumes, five times our original dataset size, and trained the classifier on this synthetic data alone. Remarkably, this approach outperformed training with the complete set of real data by 8\%. We then evaluated the model's performance on an external dataset and with unseen prompts in a zero-shot scenario, proving GenerateCT's high generalization.

GenerateCT synthesizes high-fidelity 3D chest CT volumes from free-form text prompts. To our knowledge, this is the first approach to explore text-to-3D medical imaging generation. Our contributions can be summarized as follows:

\begin{itemize}
\item We propose a novel text-to-CT generation framework capable of producing 3D chest CT volumes conditioned on medical language text prompts.
\item At the core of GenerateCT is CT-ViT, which enables autoregressive encoding and decoding of 3D CT volumes for flexible axial field-of-view handling.
\item We conduct a thorough evaluation of our approach's generative abilities compared with reasonably designed baselines across multiple image-quality metrics. Also, human domain experts evaluate generated 3D chest CT volumes underscoring a high degree of text alignment, realism, and consistency.
\item We assess the generated volumes' clinical value and text-alignment by performing a multi-abnormality classification task in two settings: (a) data augmentation, with an increase of up to a factor of five, and (b) a zero-shot scenario, where no prompts from the training set are used for generation.
\item To facilitate out-of-the-box 3D chest CT volume generation based on text prompts, we make all trained models (and code) publicly available.
\end{itemize}

\begin{figure}
    \centering
    \includegraphics[width=1\textwidth]{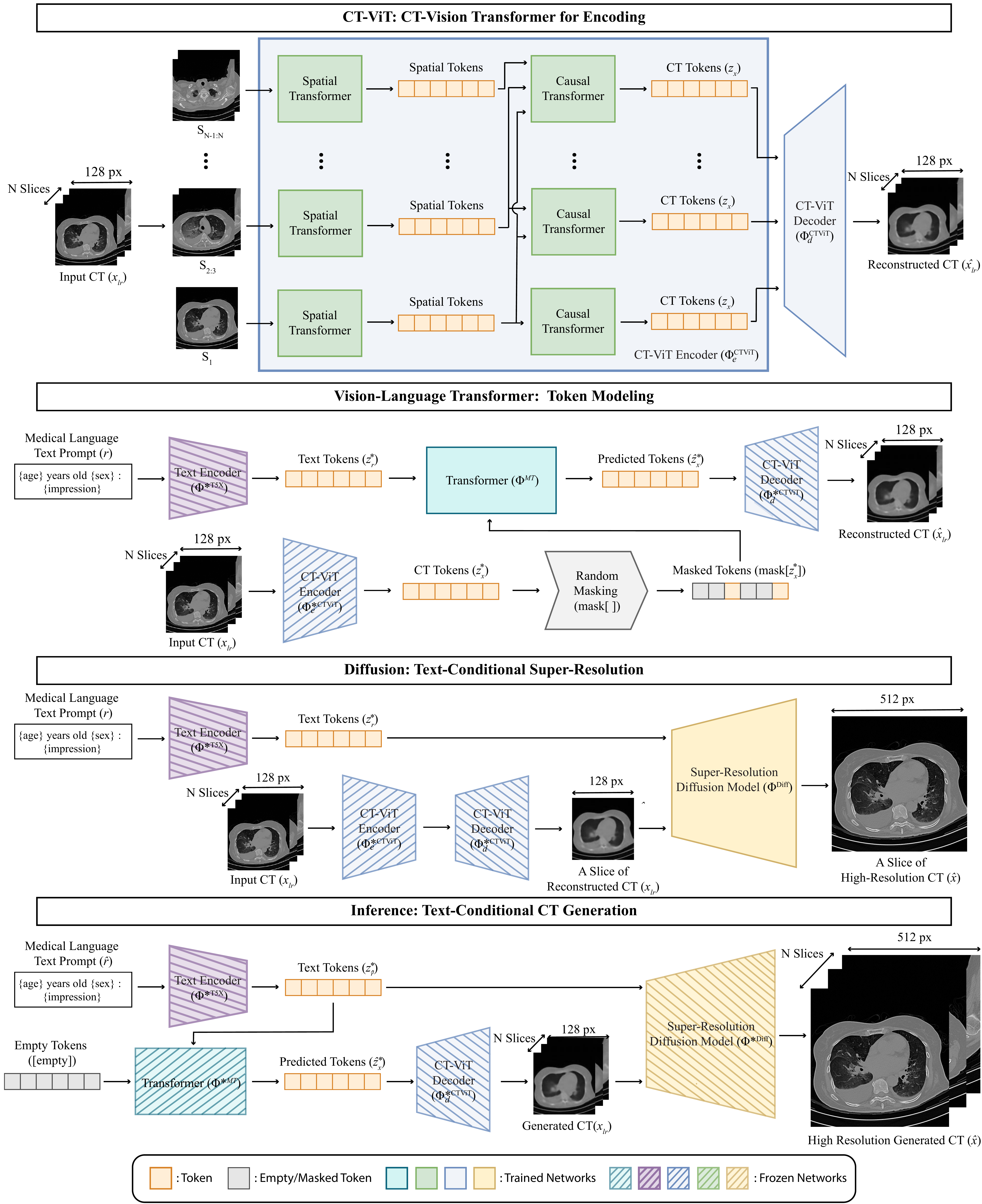}
    \caption{The GenerateCT architecture consists of three main components. (1) The CT-ViT encoder architecture processes the embeddings of CT patches from raw slices S through a spatial transformer followed by a causal transformer (auto-regressive in-depth), generating CT tokens. (2) The vision-language transformer is trained to reconstruct masked tokens based on the frozen CT-ViT encoder's predictions, conditioned on T5X text prompt tokens. (3) A text-conditional diffusion model is employed to upsample low-resolution slices from generated 3D chest CT volumes. Finally, GenerateCT demonstrates the capability to generate high-resolution 3D chest CT volumes with arbitrary slice numbers conditioned on medical language text prompts.}
    \label{framework}
\end{figure}

\section{Related Works}
\label{sec:related}

\textbf{Text-conditioned medical image generation.}
Due to the increasing demand for data, medical image generation has emerged as an important research direction. Recent studies~\cite{chambon2022roentgen, lee2023unified} have explored the generation of 2D medical images based on medical language text prompts. These studies have successfully adapted pre-trained latent diffusion models~\cite{rombach2022high}, utilizing publicly available chest X-rays and corresponding radiology reports~\cite{johnson2019mimic}. With GenerateCT, we expand this capability to include the text-conditioned generation of 3D medical imaging.

\noindent\textbf{Text-conditioned video generation.}
has seen significant advancements and can be split into two primary research streams: diffusion-based~\cite{ho2022imagen,blattmann2023align,yang2022diffusion,voleti2022masked,ho2022video} and transformer-based auto-regressive methods~\cite{villegas2022phenaki,wu2022nuwa,hong2022cogvideo,wu2021godiva}. Diffusion-based techniques, utilizing 3D U-Net architectures, typically generate shorter and low-resolution videos with a preset number of frames, but can enhance resolution and duration through cascaded diffusion models~\cite{ho2022imagen}. In contrast, transformer-based methods offer adaptability, handling variable frame numbers, and producing longer videos, albeit at lower dimensions~\cite{yan2021videogpt}. In this context, our method extends the concept of text-conditional video generation to 3D medical imaging, essentially treating 3D CT volumes as a video. GenerateCT combines a transformer-based~\cite{villegas2022phenaki} and a diffusion-based method~\cite{ho2022imagen}, which enables the generation of high-resolution CT volumes with flexible and increased slice counts.

\noindent\textbf{Datasets for text-conditioned medical image generation.} 
Training models to generate medical images from text requires paired imaging data with corresponding radiology reports. While publicly available 2D medical imaging datasets like MIMIC-CXR~\cite{johnson2019mimic} exist, there is a scarcity of publicly accessible 3D medical imaging datasets with reports. Creating such datasets is challenging due to their larger size, the expertise required for annotating 3D images, and strict data-sharing restrictions. The limited availability of such datasets is evident, as even a study focusing on multi-abnormality detection in chest CT volumes~\cite{draelos2021machine} made only a small portion of their dataset publicly accessible. This highlights the urgent need for more publicly available 3D medical imaging data and the potential for text-conditioned 3D medical image data generation, which can drive further research in this field. To address this challenge, we have made our fully trained models publicly available. We hope that this will enable researchers to generate their own datasets using text prompts, virtually without restrictions.

\section{Method}

\subsection{Dataset Preparation}
\label{dataset}
Our study received ethical approval from the Clinical Research Ethics Committee at Istanbul Medipol University (E-10840098-772.02-6841, 27/10/2023). We utilize chest CT volumes and corresponding radiology reports from the CT-RATE dataset, which is rigorously anonymized to uphold patient privacy \cite{hamamci2024foundation}. Our training data comprises 25,701 CTs with a $512 \times 512$ resolution and varying slice counts ranging from 100 to 600. These volumes originate from 21,314 unique patients and are reconstructed using multiple methods appropriate for different window settings \cite{willemink2019evolution}, resulting in 49,138 CT volumes. We divided the dataset into a training set comprising 20,000 unique patients and a testing set comprising 1,314 unique patients, ensuring no patient overlap. Each CT is accompanied by metadata that includes the patient's age, sex, and imaging specifics. Moreover, these volumes are paired with radiological reports that are categorized into separate sections: clinical information, technique, findings, and impression. The prompts are formatted as \textit{\{age\}~years~old~\{sex\}:~\{impression\}} using the impression section and metadata, as shown in~\cref{first}. We convert the CTs into their respective Hounsfield Units (HU) using the slope and intercept values retrieved from the metadata. These values are clipped to the range $[-1000~\text{HU}, +1000~\text{HU}]$, representing the practical lower and upper limits of the HU scale \cite{denotter2019hounsfield, lamba2014ct}.

\subsection{GenerateCT: Text-Conditional 3D CT Generation}
GenerateCT, as shown in~\cref{framework}, consists of three primary components, each trained in distinct stages: (1) CT-ViT for 3D CT volume encoding, (2) a masked generative image-text transformer for text and image alignment, and (3) text-conditional diffusion models for super-resolution. It processes a 3D CT volume, $x \in \mathbb{R}^{Z \times H \times W}$, covering sagittal ($H$), coronal ($W$), and axial ($Z$) dimensions, alongside a corresponding text report, $r$. GenerateCT is trained to create 3D CT volumes from medical text prompts, with dimensions set to $H=512$, $W=512$, and $Z=201$ in our experiments. Below, each component is explained in detail.

\noindent\textbf{CT-ViT: 3D CT-Vision Transformer.}\label{sec:CT-ViT} We introduce CT-ViT to achieve compact latent representations of 3D volumes. Inspired by video transformers like ViViT~\cite{arnab2021vivit} and C-ViViT~\cite{villegas2022phenaki}, CT-ViT extracts spatiotemporal tokens from the CT volume. These tokens are encoded through both all-to-all spatial attention and causal attention layers, resulting in encoded CT tokens. Subsequently, a decoder network operates on these tokens to recreate the input 3D CT volume, forming an autoregressive encoder-decoder network. This design is advantageous for handling real-world 3D CT volumes with variable cranio-caudal coverage.

The CT-ViT encoder network ($\Phi^\text{CTViT}_e$) accepts a low-resolution CT volume $x_{lr}\in{\mathbb{R}^{(201)\times 128\times 128}}$ and outputs embedded CT tokens $z_{x}\in{\mathbb{R}^{(101)\times 8\times 8}}$. The decoder network ($\Phi^\text{CTViT}_d$) then utilizes these embedded CT tokens to reconstruct CT volumes ($\hat{x}_{lr}$) in the same space. Concisely, the process is represented as:
$$
z_{x} = \Phi^{\text{CTViT}}_{e}(x_{lr}) \text{~~and~~} \hat{x}_{lr} = \Phi^{\text{CTViT}}_{d} ( z_{x}).
$$
The encoder network first extracts non-overlapping patches of \( 16 \times 16 \) pixels from the first slice of a 3D chest CT volume, and \( 2 \times 16 \times 16 \) patches from the remaining slices. Each patch is then linearly transformed into a \( D \)-dimensional space, where \( D \) is the latent space dimension, set to $512$. For the first frame, that data is reshaped from \( B \times C \times 1 \times (H \cdot p_1) \times (W \cdot p_2) \) to \( B \times 1 \times H \times W \times (C \cdot p_1 \cdot p_2) \). Here, \( B \) represents the batch size, \( C \) the number of channels, \( H \) and \( W \) the height and width of the slices, and \( p_1 \) and \( p_2 \) the spatial patch sizes. A linear layer then transforms the final dimension to \( D \), resulting in a tensor with dimensions \( B \times 1 \times \frac{H}{p_1} \times \frac{W}{p_2} \times D \). The remaining slices undergo a similar reshaping and linear transformation, from \( B \times 1 \times (T \cdot p_t) \times (H \cdot p_1) \times (W \cdot p_2) \) to \( B \times T \times H \times W \times (C \cdot p_t \cdot p_1 \cdot p_2) \) and finally to \( B \times T \times \frac{H}{p_1} \times \frac{W}{p_2} \times D \), with \( p_t \) representing the temporal patch size and \( T \) the number of temporal patches.

After combining the initial and subsequent frame embeddings, the resulting tensor is \( B \times (1 + T) \times \frac{H}{p_1} \times \frac{W}{p_2} \times D \). This tensor is processed by two transformer networks in sequence. The spatial transformer operates on a reshaped tensor of \( (B \cdot (1 + T)) \times (\frac{H}{p_1} \cdot \frac{W}{p_2}) \times D \), outputting a tensor of the same size. The causal transformer then processes this output, reshaped to \( (\frac{H}{p_1} \cdot \frac{W}{p_2}) \times (B \cdot (1 + T)) \times D \), and produces an output maintaining these dimensions. This process preserves both the spatial and latent dimensions after each transformer layer, ensuring 3D volumetric information retention throughout the network's processing stages.

The CT-ViT decoder mirrors the encoding process by transforming tokens back into their original voxel space, reconstructing 3D CT volumes while preserving the axial dimensionality of the input. This capability enables the generation of 3D CT volumes with varying numbers of axial slices. Additionally, CT-ViT incorporates vector quantization to create a discrete latent space. This technique quantizes the encoder outputs into a set of entries from a learned codebook, as described in \cite{van2017neural}. Besides, the model's autoregressive training process combines multiple loss functions, including the L2 loss from ViT-VQGAN \cite{vtqgan} to ensure consistency during the reconstruction, image perceptual loss \cite{perceptualimg} for perceptual similarity, and an adversarial loss function in alignment with StyleGAN~\cite{karras2020analyzing}.

\noindent\textbf{Vision-Language Transformer: Token Modeling.}\label{sec:maskgit}
In GenerateCT's second stage, we align CT and text spaces using masked visual token modeling~\cite{chang2022maskgit}. This involves the previously trained CT-ViT encoder ($\Phi^{*{\text{CTViT}}}_{e}$) and its produced CT tokens (${z^*_x}$), which are masked ($\mathtt{mask}[z^*_x]$) and input into a bidirectional transformer ($\Phi^{MT}$). The radiology report ($r$), encoded with a text encoder ($\Phi^\text{T5X}$), serves as a conditional input~\cite{raffel2020exploring}. The transformer's role is to predict these masked CT tokens based on the text embedding, incorporating cross-attention with the input CT tokens. These predicted CT tokens are then processed by the frozen CT-ViT decoder ($\Phi^{*{\text{CTViT}}}_{d}$), expected to reconstruct the input 3D CT volume accurately. The forward pass in the text-CT alignment stage, utilizing masked token modeling with the trained CT-ViT, is represented as follows:
$$
\hat{z}^*_x = \Phi^{MT}\big(\mathtt{mask}[{z^*_x}], \Phi^{*\text{T5X}}(r)\big) \text{~~and~~}\hat{x}_{lr} = \Phi^{*{\text{CTViT}}}_{d}(\hat{z}^*_x).
$$
The training for this vision-language transformer model also integrates reconstruction loss and token critic loss. Reconstruction loss assesses the model's capability to predict masked video codebook IDs in sequences, using cross-entropy to quantify the difference between predicted and actual tokens. Additionally, the critic loss includes a component evaluating whether video codebook ID sequences are authentic or fabricated, employing binary cross-entropy to gauge the alignment between the predicted critics' probabilities and the actual labels.

During inference, all CT tokens are masked and predicted by the bidirectional transformer, based on the text embeddings and the CT tokens previously predicted. These tokens are then reconstructed using the CT-ViT decoder.

\noindent\textbf{Diffusion Steps: Text-Conditional Super-Resolution.}\label{diffusion} 
GenerateCT's final stage employs a diffusion-based, text-conditional super-resolution model ($\Phi^\text{Diff}$) to enhance the resolution of each slice from initially synthesized low-resolution 3D CT volumes in the axial dimension. Using a cascaded diffusion approach~\cite{ho2022cascaded}, this process sequentially employs diffusion steps that enhance image resolution by upsampling and introducing finer details. This cascaded method outperforms traditional U-Net diffusion models~\cite{saharia2022palette, hamamci2023diffusion} in terms of memory efficiency, achieved by incorporating a cross-attention layer with T5 embedded text tokens at the bottleneck stage, which replaces self-attention layers~\cite{beltagy2020longformer}. This layer conditions the diffusion on both the encoded text prompt and the initial low-resolution image. Optimal cascading steps have been identified through an ablation study (\cref{table}). Notably, using CT-ViT reconstructed volumes as input, instead of the original downsampled volumes, enhances performance, aligning with the principles of noisy conditioning~\cite{ho2022imagen}. The upsampling process is denoted as $ \hat{x} = \Phi^\text{Diff}\big(\hat{x}_{lr}, \Phi^{*\text{T5X}}({r})\big),$ where $\hat{x}$ represents the final generated high-resolution 3D chest CT volume, with dimensions of $(201) \times 512 \times 512$, based on the prompt.

The training of the model employs a loss function designed to minimize the disparity between denoised and actual high-resolution images. This function incorporates a Mean Squared Error (MSE) component for pixel accuracy and integrates noise levels into the loss weighting, ensuring that noisy samples are properly accounted for. The overall loss is the mean of these noise-weighted MSE values, quantifying the denoised slices' deviation from the actual slices.

\noindent\textbf{Inference.}\label{inference}
After training, GenerateCT can generate 3D chest CT volumes ($\tilde{x}$) from a given novel radiological text prompt ($\tilde{r}$), formally defined as follows:
$$
z^*_r = \Phi^{*\text{T5X}}(\tilde{r}) {~~\text{and~~}} \tilde{x} = \Phi^{*\text{Diff}}\big(\Phi^{*{\text{CTViT}}}_{d}~\Phi^{*MT}([\mathtt{empty}], z^*_r ), z^*_r\big), 
$$
where [empty] represents fully masked CT token placeholders. This process involves encoding the prompt, predicting CT tokens with the masked transformer, and then decoding these tokens to create the synthetic 3D CT volume.

\subsection{Implementation Details}
We trained the CT-ViT model on 49,138 CT volumes (see \cref{dataset}). We employed the Adam optimizer~\cite{kingma2014adam} with $\beta$1 and $\beta$2 hyperparameters set to 0.9 and 0.99, respectively, a learning rate of 0.00003, and an effective batch size of 32. The training was conducted for one week on a node with 8 A100 GPUs, completing 100,000 iterations. Subsequently, we trained the MaskGIT transformer using a paired dataset, which included the same 3D CT volumes with the same resolution as CT-ViT and medical language text prompts~(\cref{framework}) from their corresponding radiology reports. The Adam optimizer was used with identical $\beta$1 and $\beta$2 values, and the learning rate was maintained at 0.00003. However, we adjusted the effective batch size to 4 and introduced a cosine annealing warmup scheduler with a warmup phase of 10,000 steps and a maximum limit of 4,000,000 steps. This training stage, also executed on 8 A100 GPUs, lasted one week, concluding after 500,000 iterations. Finally, we trained the super-resolution diffusion model on the CT slices, each initially resized to $128 \times 128$. The super-resolution model then upscaled these to $512 \times 512$, using the original volumes as ground truth. For this model, the same text prompts used for the 3D volumes were provided as conditioning for all slices of a 3D CT. We retained the previous hyperparameters for the Adam optimizer and set the learning rate to 0.0005. This final training phase was carried out on 8 A100 GPUs over a week, reaching 275,000 iterations.

\section{Experimental Results}

\definecolor{mycolor}{rgb}{0.90, 0.90, 0.90} 

\begin{table}[t]
  \centering

  \caption{Quantitative results for GenerateCT and its variants, compared with baseline methods, demonstrate our method's superior performance across all key metrics, underscoring its effectiveness in generating 3D chest CT volumes from medical text prompts. Sampling time tests were conducted on an NVIDIA A100 80GB GPU.}
  \setlength{\tabcolsep}{5pt} 
  \renewcommand{\arraystretch}{0.3} 
  \begin{tabular}{
    @{}
    l
    c
    c
    S[table-format=4.1]
    S[table-format=1.4]
    S[table-format=3.1]
    S[table-format=2.1]
    @{}
  }
    \toprule
    {\textbf{Method}} & {\textbf{Out}} & {\textbf{Time(s)}} & {\textbf{FVD\(_{\text{I3D}}\)\(\downarrow\)}} & {\textbf{FVD\(_{\text{CT-Net}}\)\(\downarrow\)}} & {\textbf{FID\(\downarrow\)}} & {\textbf{CLIP\(\uparrow\)}} \\
    \midrule
    Base w/ Imagen & 2D & 234 & {3557.7} & {17.319} & {160.8} & {24.8} \\ 
        Base w/ SD & 2D & 367 & {3513.5} & {21.194} & {151.7} & {23.5} \\
    Base w/ Phenaki & 3D & 23 & 1886.8 & 9.5534 & 104.3 & 25.2 \\ 
    \midrule
    Ours (2SCM) & 3D & 102 & 1661.4 & 8.9021 & 86.9 & 25.9 \\
    \rowcolor{mycolor} 
    Ours (3SCM) & 3D & 184 & \textbf{1092.3} & \textbf{8.1745} & \textbf{ 55.8} & \textbf{27.1} \\
    Ours (4SCM) & 3D & 244 & 1201.4 & 8.5869 & 71.3 & 26.6 \\
    \bottomrule
  \end{tabular}

  \label{table}
\end{table}

\noindent\textbf{Quantitative evaluation.}
We evaluated the quality of 3D chest CT volumes generated by different methods utilizing the following metrics, see~\cref{table}:
\begin{itemize}
    \item 
Fréchet Video Distance (FVD) quantifies the dissimilarity between generated and real volumes by extracting image features with the I3D model~\cite{unterthiner2019fvd}, which is well-suited for videos, denoted as $\text{FVD}_{\text{I3D}}$. Recognizing its limitations for medical imaging, we also employed the CT-Net model~\cite{draelos2021machine}, trained on our dataset (detailed in~\cref{section4.3}). This approach ($\text{FVD}_{\text{CT-Net}}$) allows for domain-relevant feature extraction, providing a more appropriate comparison.
    
    \item 
Fréchet Inception Distance (FID) assesses the quality of generated images, but at a slice-level using the InceptionV3 model~\cite{heusel2017gans}. FID may not be fully suitable for our 3D generation, as individual 2D CT slices might not accurately reflect volume-level findings, potentially leading to misleading results.
    \item 
The CLIP score quantifies the alignment between text prompts and generated volumes, a process achieved by utilizing the pretrained CLIP model~\cite{radford2021learning}, which is designed to correlate visual and textual content effectively. Within our training dataset, comprising paired volumes and radiology text reports, we attained a CLIP score of $\mathbf{27.4}$, serving as a benchmark for alignment.

\begin{figure}[t]
    \centering
    \includegraphics[width=1\linewidth]{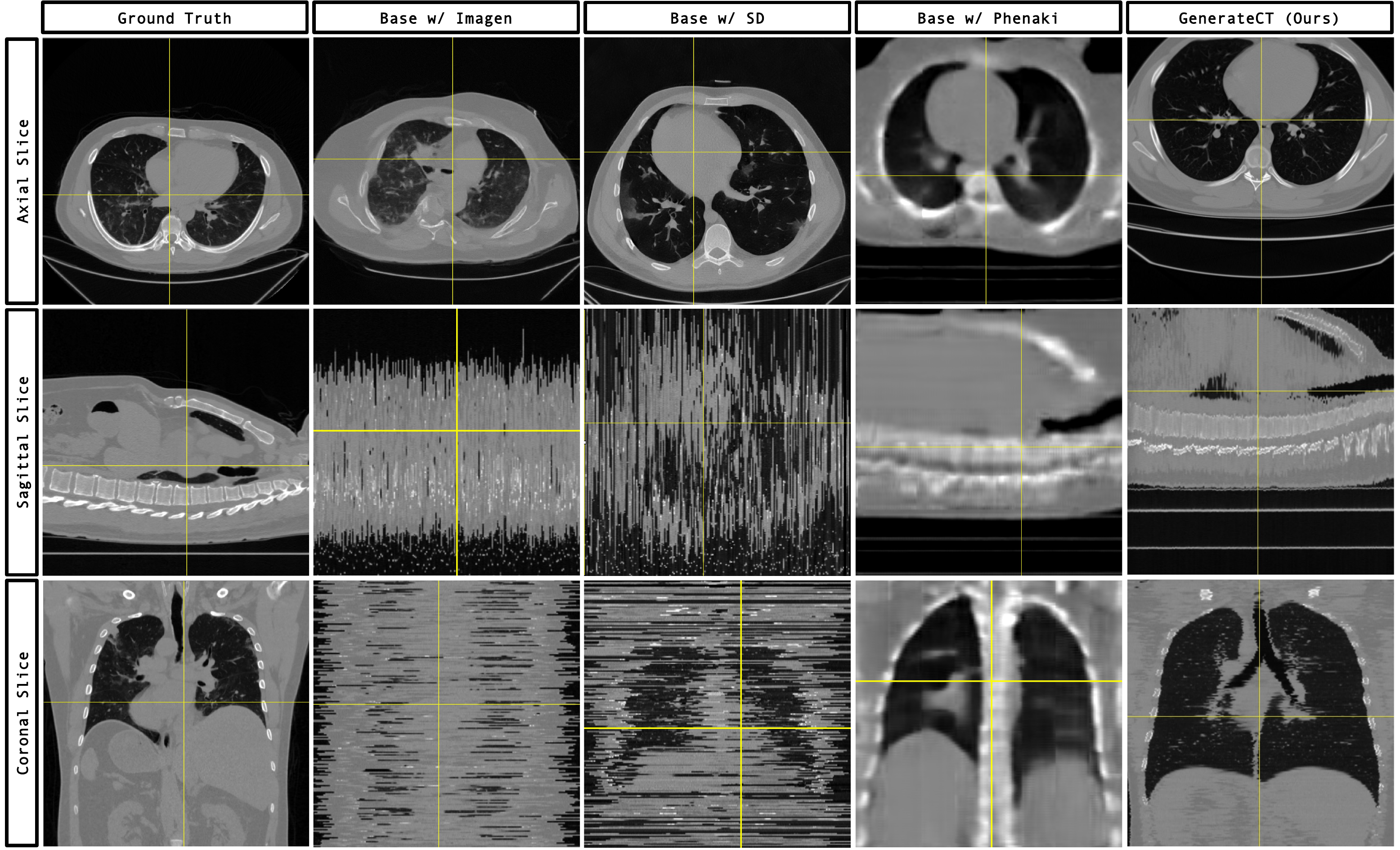}
    \caption{Axial, sagittal, and coronal slices of 3D CT volumes generated by various methods based on the text prompt: \emph{"26 years old male: Findings compatible with COVID-19 pneumonia".} The results highlight GenerateCT's proficiency in creating detailed, spatially consistent 3D CTs. Comparing with ground truth, though uncommon in text-to-image research, serves as a reference here, showcasing GenerateCT's ability to produce diverse CTs accurately aligned with prompts, rather than just replicating training data.}
    \label{planes}

\end{figure}

\end{itemize}

\noindent\textbf{Comparisons with baseline methods.} Given GenerateCT's uniqueness as the first framework of its kind in 3D medical imaging, there are no directly comparable methods. Thus, to demonstrate its effectiveness, we designed the following baseline methods for comparative analysis (see~\cref{table}), each selected to highlight different facets of GenerateCT’s innovative solution for creating clinically accurate and consistent 3D chest CT volumes from text prompts.
\begin{itemize}
    \item \textbf{Base w/ Imagen.} To assess the importance of our 3D generation architecture for achieving spatial consistency in 3D chest CT volumes, we employed a text-conditional 2D image generation method, Imagen \cite{saharia2022photorealistic}, for slice-wise generation. We conditioned Imagen on the slice number alongside the text prompt during training. Then the generated slices are combined in the order of the conditioning slice number to form a 3D chest CT volume. \cref{planes} shows that, even though high resolution and accurate axial slices were achieved, the chest CT volumes generated by this 2D baseline lack spatial consistency, further highlighting the need for a dedicated 3D generation algorithm.
    \item \textbf{Base w/ SD.} To demonstrate that even a pre-trained 2D text-to-image model is not sufficient for 3D medical image generation, we fine-tuned Stable Diffusion (SD) \cite{rombach2022high}. Despite slightly outperforming Imagen, fine-tuning SD failed to produce spatially consistent and accurate 3D volumes, as seen in the sagittal and coronal planes (\cref{planes}). This effort also highlighted \emph{the computational complexities of direct, text-conditional 3D medical image generation}: generating just one 2D axial slice with SD required 13 GB of GPU memory. The memory requirement would escalate exponentially when utilizing a basic 3D diffusion model to generate a 3D chest CT volume consisting of over 200 slices, underscoring its limitations for such medical applications and the imperative for an optimally engineered framework like GenerateCT.
    \item \textbf{Base w/ Phenaki.} To highlight that even 3D generation models might not capture the nuanced medical details of chest CT volumes, we adapted a state-of-the-art text-to-video generation model, Phenaki \cite{villegas2022phenaki}, for 3D chest CT generation. Although spatial consistency increased, Phenaki failed to generate medically detailed CT volumes (\cref{planes}). This underscores the unique challenges of text-conditional high-resolution 3D medical image generation and the necessity for an optimized solution like our cascaded architecture.
\end{itemize}

\begin{figure}[t]
    \centering
    \includegraphics[width=1\textwidth]{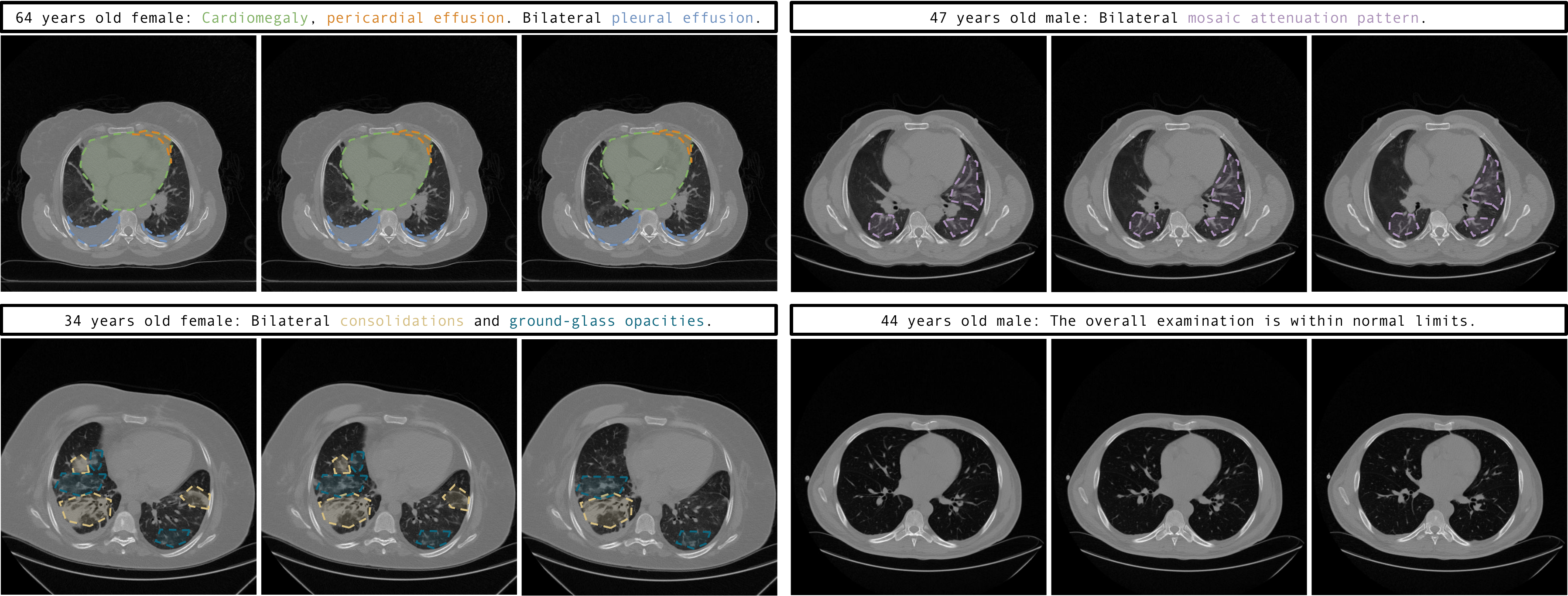}
    \caption{Three sequential slices from each synthetic 3D chest CT within the practical HU range of $[-1000~\text{HU}, +1000~\text{HU}]$ generated based on the given prompt, showcasing GenerateCT's proficiency in preserving spatial consistency across successive slices. Abnormalities referenced in the prompts are color-highlighted, underscoring our method's precision in translating textual descriptions into clinically accurate volumetric features.}
    \label{result}
\end{figure} 

\noindent\textbf{Ablation study.} GenerateCT's cascaded architecture was evaluated across different stages. We tested three $X$-Stage Cascaded Models ($X$SCM), which combine a transformer-based 3D text conditional generation model, followed by $X-$1 diffusion-based super-resolution steps to produce high-resolution 3D CT volumes. As seen in \cref{table}, $\text{FVD}_{\text{CT-Net}}$ consistently showed lower scores compared to $\text{FVD}_{\text{I3D}}$, a result of CT-Net's specific training on 3D CT volumes. As the number of super-resolution steps increased, both $\text{FVD}_{\text{I3D}}$ and $\text{FVD}_{\text{CT-Net}}$ along with FID and CLIP showed enhanced performance. However, the 4SCM model was an outlier due to its significantly low initial resolution. The 3SCM model, achieving a CLIP score of $27.1$ close to the baseline of $27.4$, demonstrated excellent alignment with the text prompts. Therefore, the 3SCM model, outperforming others in all key metrics, was selected as the optimal configuration for GenerateCT.

\begin{figure}[ht]
    \centering
    \includegraphics[width=1\linewidth]{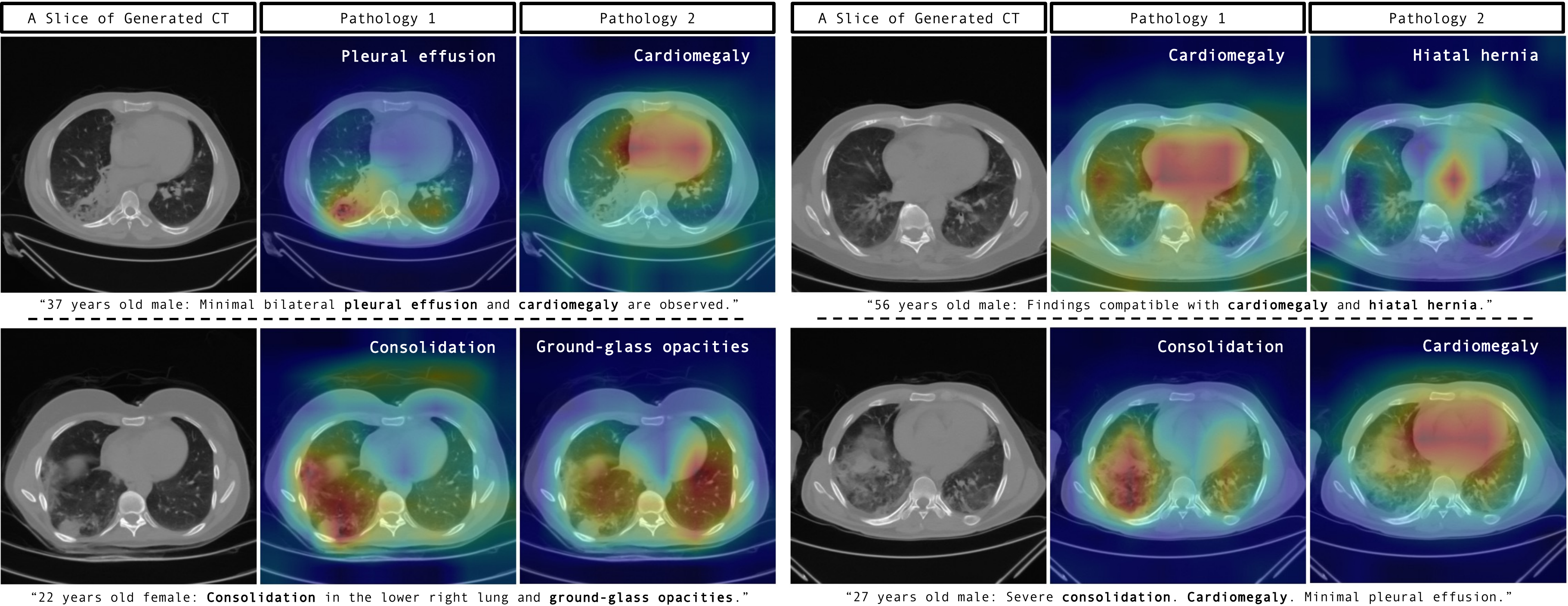}
    \caption{Cross-attention maps for showing specific abnormalities in the text-conditional generation of chest CT volumes, highlighting GenerateCT's precision in aligning text with relevant regions. Colors from blue to red represent the weights from low to high.}
    \label{attention}
\end{figure}

\noindent\textbf{Qualitative results.} GenerateCT effectively translates specific text prompts into 3D chest CT volumes, as shown in \cref{result}. The initial three volumes show distinct pathologies, marked with colored text and areas, consistent across slices, contrasting with a fourth volume of a healthy lung. These volumes display diversity in size, orientation, age, and sex, emphasizing the range of data producible from the text prompts. \cref{planes} further demonstrates GenerateCT's ability to create comprehensive 3D images by including both sagittal and coronal slices in addition to axial ones. \cref{attention} showcases the model's cross-attention between text and generated volumes, emphasizing regions corresponding to specific pathologies. This involves averaging attention outputs across heads and relevant tokens corresponding to each pathology in the input prompts, then upscaling the low-dimensional cross-attention outputs to high-resolution CT volume dimensions using an affine transformation. Such visualizations show GenerateCT's precision in aligning text with the relevant regions, translating medical terms into spatially accurate and clinically significant image features, such as cardiomegaly around the heart, pleural effusion at the effusion site, and consolidation in the affected lung area. We showcase slices from 3D chest CT volumes in the raw HU range of $[-1000, +1000]$, diverging from standard windowing for more authentic representation. Supplementary material offers varied windowing examples.

\begin{table}[t]
\caption{Labeling outcomes by experts for authenticity prediction and text prompt alignment with real and synthetic 3D CT volume. The statistical analysis underscores the convincing realism and text alignment of the generated 3D chest CT volumes.}
\centering
\small
\begin{tabular}{cccccc}
\toprule
 & \multicolumn{2}{c}{First Radiologist (4 years)} & \multicolumn{2}{c}{Second Radiologist (11 years)} \\
\cmidrule(lr){2-3} \cmidrule(lr){4-5}
Task & Real Volumes & Synthetic Volumes & Real Volumes & Synthetic Volumes \\
\midrule
 & \cellcolor{Blue74}\textcolor{white}{Real: 74} & \cellcolor{Blue41}Real: 41 & \cellcolor{Blue71}\textcolor{white}{Real: 71} & \cellcolor{Blue36}Real: 36 \\
\multirow{-2}{*}{3D Realism} & \cellcolor{Blue26}Synthetic: 26 & \cellcolor{Blue59}\textcolor{white}{Synthetic: 59} & \cellcolor{Blue29}Synthetic: 29 & \cellcolor{Blue64}\textcolor{white}{Synthetic: 64} \\
\addlinespace
 & \cellcolor{Blue82}\textcolor{white}{Matched: 82} & \cellcolor{Blue66}\textcolor{white}{Matched: 66} & \cellcolor{Blue83}\textcolor{white}{Matched: 83} & \cellcolor{Blue70}\textcolor{white}{Matched: 70} \\
\multirow{-2}{*}{Alignment} & \cellcolor{Blue18}Mismatched: 18 & \cellcolor{Blue34}Mismatched: 34 & \cellcolor{Blue17}Mismatched: 17 & \cellcolor{Blue30}Mismatched: 30 \\
\bottomrule
\end{tabular}
\label{expert}
\end{table}

\noindent\textbf{Expert evaluation.}
A blinded study with two radiologists (4 and 11 years of experience) evaluated 200 3D chest CT volumes (100 real, 100 synthetic). The experts were tasked with determining whether each volume was real or synthetic and with verifying the match between text prompts and volume findings (\cref{expert}). In the first task, even though they were aware that half of the volumes were synthetic and that 3D volumes were provided for evaluation, both radiologists exhibited significant misclassification rates. This underscores GenerateCT's ability to create indistinguishable and spatially accurate 3D CT volumes. The disparity in false negative rates for real versus synthetic volumes was not statistically significant ($p=0.0636$, unpaired T-test), emphasizing synthesized volumes' realism. In the second task, the radiologists found that a comparable number of synthetic volumes, such as $70$, accurately matched the given text prompts, similar to the real volumes. This indicates a high level of alignment between the generated 3D chest CT volumes and their corresponding text prompts.

\begin{figure}[t]
    \centering
    \includegraphics[width=1\linewidth]{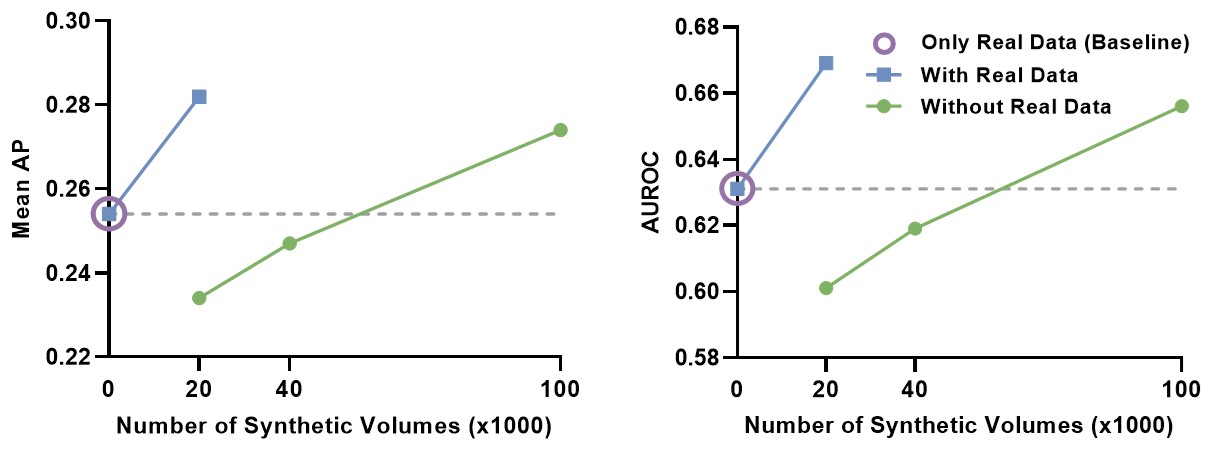}
    \caption{A comparative analysis of multi-abnormality classification models with incremental data augmentation using GenerateCT highlights significant clinical utility, in low-data environments. The mean frequency of abnormalities in the test set is $0.179$.}
    \label{data_aug}
\end{figure}

\section{Clinical Value of GenerateCT}

\noindent\textbf{Utilizing GenerateCT in data augmentation.}\label{section4.3}
We assessed the clinical potential of GenerateCT within a radiological framework. To set a benchmark, a multi-abnormality classification model \cite{draelos2021machine} was initially trained on 20,000 real 3D chest CT volumes from our dataset, each representing a unique patient profile (see \cref{dataset}). The baseline achieved a mean average precision (AP) of $0.254$ and an area under the receiver operating characteristic curve (AUROC) of $0.631$. We then generated 20,000 synthetic volumes using text prompts and trained the classifier on this mixed dataset of real and synthetic data. The results showed an $11\%$ improvement in mean AP and a $6\%$ increase in mean AUROC compared to training on real data alone (see \cref{data_aug}). Further experimentation involved expanding the synthetic dataset to 100,000 3D volumes using repeated prompts. Training exclusively on this synthetic data led to an $8\%$ increase in mean AP and a $4\%$ rise in mean AUROC compared to the real-data model. Given the synthetic-data model’s outperformance over the real-data model, alongside computational limitations (each generated volume takes 184 seconds and is 400 MB, totaling 40 TB for 100,000 volumes), further extensions have not been pursued.

The results, detailed in \cref{data_aug}, demonstrate GenerateCT's effectiveness in clinical settings. First, data augmentation, even by a single factor, significantly boosts performance, underscoring its potential for researchers who have real-world data and aim to enhance performance. Second, training on a larger, fully synthetic dataset after a fivefold increase yielded notably better scores compared to the real-data-only model, highlighting GenerateCT's contribution to data privacy. This approach enables researchers to train and share generation models, like ours, facilitating the creation of synthetic data using text prompts, thus having even better performance without privacy or data-sharing concerns. Third, the increase in scores with repetitive prompt use for data generation indicates GenerateCT's ability to generate variable data using the same prompts. Further training details and accuracies by abnormality type are in the supplementary.

\noindent\textbf{Utilizing GenerateCT in a zero-shot setting.} \label{section4.4}
To evaluate GenerateCT's ability to generalize to external datasets, we conducted an experiment using RadChestCT~\cite{draelos2021machine}, which consists of 3,630 chest CT volumes with a mean abnormality label frequency of $0.129$. We created a new dataset using text prompts not included in our GenerateCT training, matching RadChestCT's training set in terms of volume count and abnormality distribution. The classifier~\cite{draelos2021machine} was trained on this synthetic dataset, the original RadChestCT dataset, and a combination of both. The results were promising: the model trained on the synthetic data achieved close performance metrics to the model trained on real patient data, with mean APs of $0.177$ (real) versus $0.146$ (synthetic) and mean AUROCs of $0.613$ (real) versus $0.536$ (synthetic). Training on the combined dataset significantly increased the performance (mean AUROC $0.623$, mean AP $0.190$). This demonstrates that GenerateCT's key benefits extend to external datasets and its potential for clinical applications, even with unseen prompts. The supplementary provides further training details and accuracies by abnormality type.

\section{Conclusion and Discussion}
In this paper, we introduce GenerateCT, the first text-conditional 3D medical image generation framework. Our experiments demonstrate its capability to generate realistic, high-quality 3D chest CT volumes from text prompts, and its clinical applications in multi-abnormality classification. We make GenerateCT fully open-source to lay a solid foundation for future research and development.

\noindent\textbf{Limitations.} Despite its innovation, GenerateCT faces several challenges. The lack of benchmarks, due to its uniqueness, limits evaluation. While it handles 3D CTs of varying sizes, a detailed assessment of this capability is needed. Our dataset, sourced from a single institution, may lack diversity, raising concerns about bias and limited applicability. Expanding training beyond the impression sections could enhance outcomes. The significant computational demands also pose challenges in resource-constrained settings. Importantly, the model may not be directly usable in real clinical settings. Although the experiments, especially in the clinical value section, might suggest readiness for real use, this is not the case. Further validation and testing in diverse clinical environments are necessary.

\noindent\textbf{Acknowledgments.} We thank the Helmut Horten Foundation for their support and Istanbul Medipol University for providing the CT-RATE dataset.

%
%
\bibliographystyle{splncs04}
\bibliography{main}

\clearpage
\setcounter{page}{1}
\setcounter{section}{0}
\setcounter{figure}{0}
\setcounter{table}{0}
\maketitlesupplementary

\begin{figure}[ht]
    \centering
    \includegraphics[width=1\linewidth]{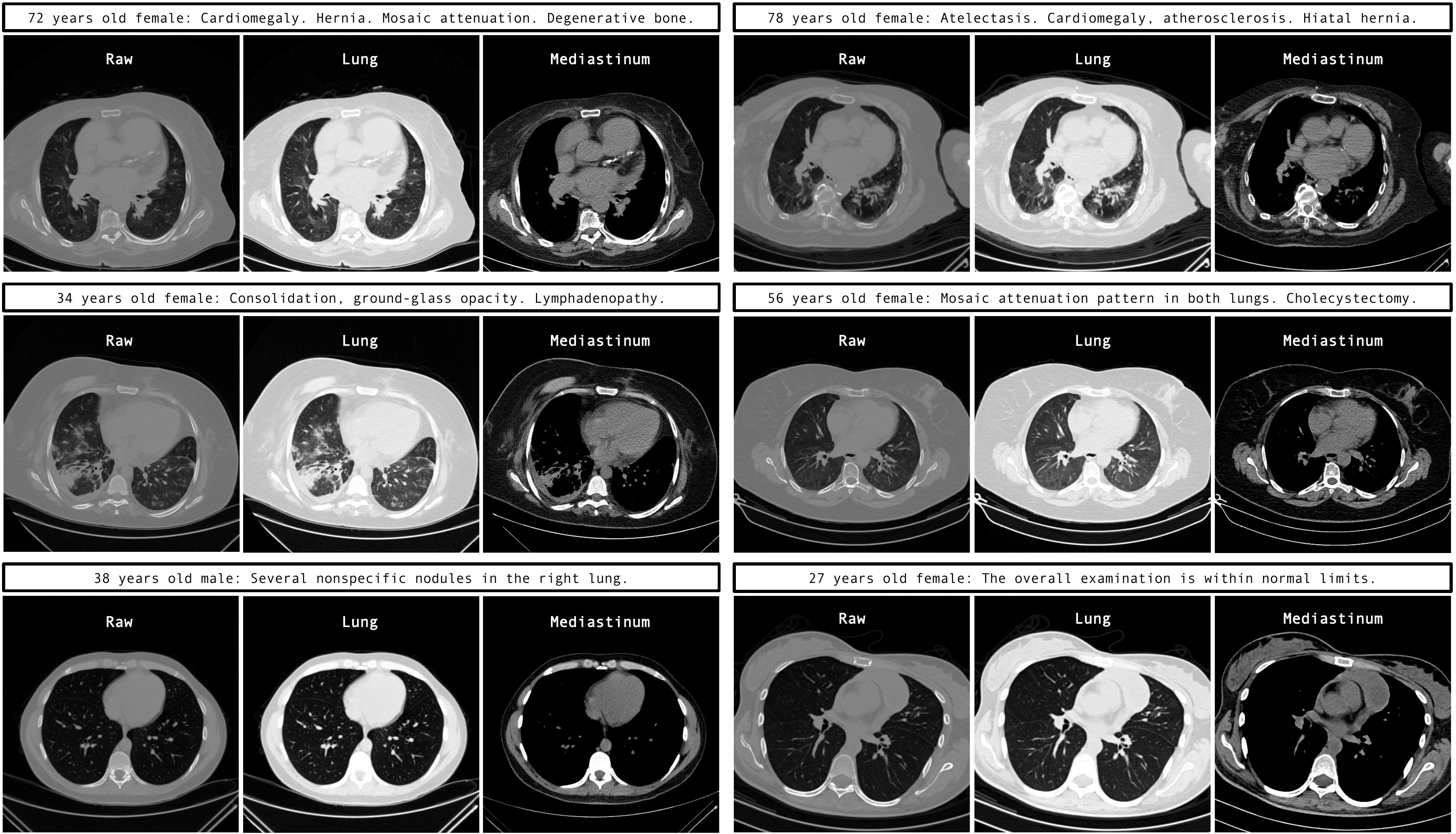}
    \caption{Example 2D slices of generated 3D CT volumes with varied windowing settings. Each example includes three windowing settings for the same slice: (1) within the raw HU range of $[-1000~\text{HU}, +1000~\text{HU}]$, (2) lung window within the range of $[-1000~\text{HU}, +150~\text{HU}]$, and (3) mediastinal window within the range of $[-125~\text{HU}, +225~\text{HU}]$. This highlights GenerateCT's ability to produce highly detailed and clinically accurate 3D chest CT volumes based on text descriptions.}
    \label{windowing}
\end{figure} 

This supplementary document enhances and expands upon the findings detailed in the main paper, focusing on three critical dimensions:
\begin{itemize}
\item \emph{Enhanced Qualitative Results:} It introduces an expanded collection of examples featuring various windowing techniques for comparison, demonstrating how GenerateCT effectively creates 3D CT volumes from text descriptions.
\item \emph{Detailing Clinical Application:} A comprehensive examination is presented on the utility of GenerateCT within a clinical setting, particularly focusing on its role in data augmentation for the classification of multiple abnormalities.
\item \emph{Generalization and Adaptability in Clinical Settings:} Further, it presents a detailed exploration of another practical clinical application of GenerateCT, where we illustrate its ability to generalize to external datasets and its proficiency in generating 3D chest CT volumes from unseen prompts.
\end{itemize}

\section{Comprehensive Qualitative Results}
\label{Qualitative}
\begin{figure}[t]
    \centering
    \includegraphics[width=1\linewidth]{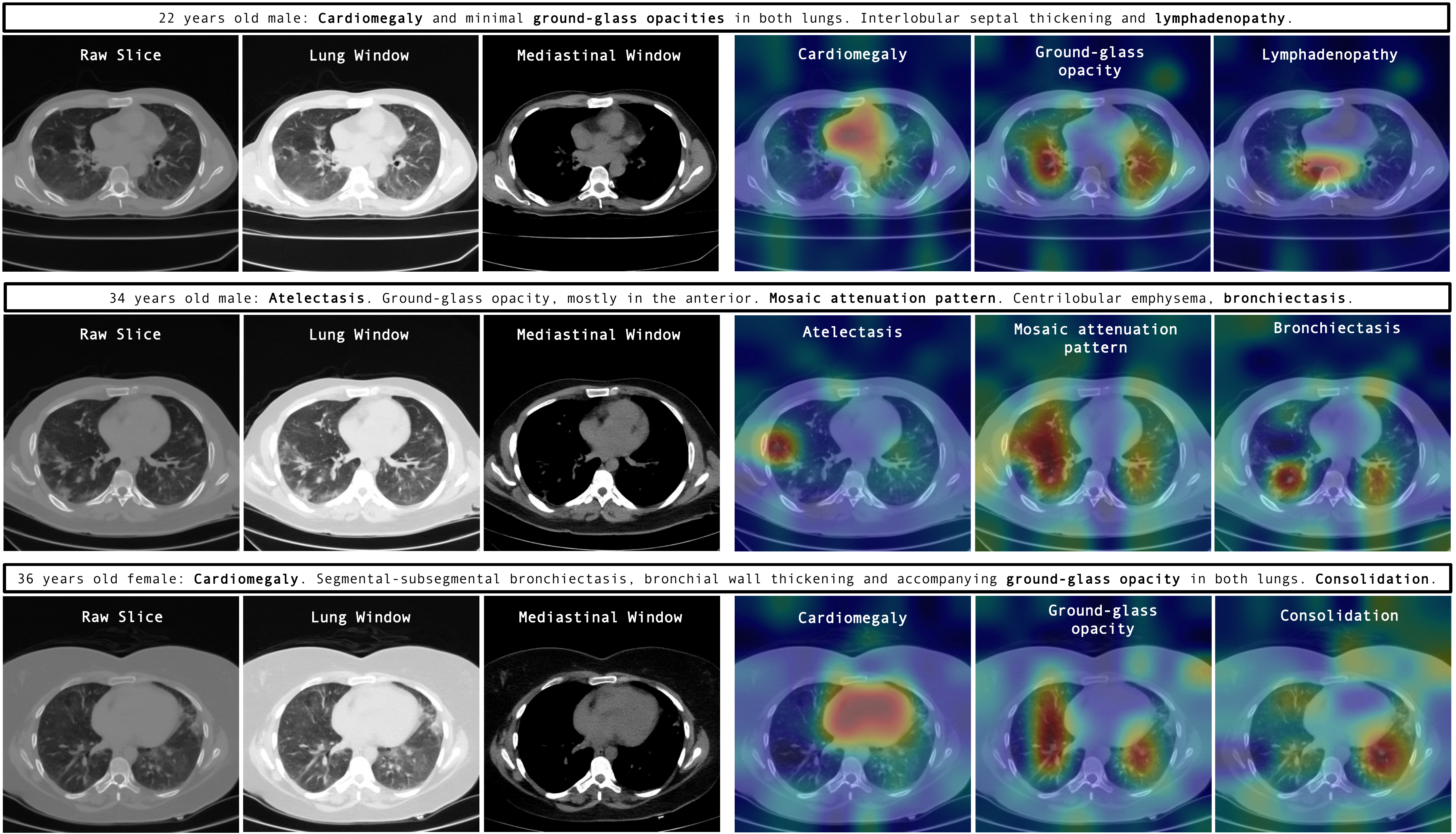}
    \caption{Cross-attention maps illustrate specific abnormalities in the text-conditional generation of 3D chest CT volumes with varied windowing settings, underscoring GenerateCT's precision in translating medical terminology into clinically relevant image features in the corresponding areas. Although our work generates comprehensive 3D chest CT volumes, we present only 2D axial slices due to presentation and visualization constraints. These slices act as representative examples to demonstrate the depth and detail GenerateCT can achieve, providing insights into its ability to accurately depict complex anatomical structures and abnormalities in a three-dimensional context.}
    \label{attention_sup}
\end{figure}

This section showcases a broad spectrum of 3D chest CT volumes generated by GenerateCT. \cref{windowing} displays 2D axial slices from synthetic 3D CT volumes, illustrating both the raw HU range of $[-1000, +1000]$ and various windowing techniques. These methods align with clinical practice and reveal the generative details derived from medical text descriptions. This emphasizes GenerateCT's precision in capturing spatial details as well as its adeptness in handling dynamic ranges. Furthermore, \cref{attention_sup} highlights the efficacy of GenerateCT's cross-attention mechanism in accurately associating specific pathologies mentioned in text prompts with the corresponding areas across different window settings. These visualizations demonstrate the model's exceptional ability to convert medical language into clinically relevant, spatially precise image features, showcasing its potential to create detailed and accurate 3D images from textual prompts.

\section{Utilizing GenerateCT in Data Augmentation}
\label{Augmentation}
In this section, we take a closer look at a practical clinical application of GenerateCT. Through a case study, we demonstrate the training of a multi-abnormality classification model with synthetic chest CT volumes generated from medical text prompts. This detailed examination underscores the substantial potential of GenerateCT in data augmentation, particularly in scenarios where obtaining real patient data is limited or challenging. Furthermore, we highlight GenerateCT's contribution to data privacy. Our approach enables researchers to train and share models similar to ours, promoting the creation of synthetic data through text prompts, thereby enhancing performance without compromising privacy or data-sharing concerns. Additionally, we show that GenerateCT can reliably generate diverse data, even when using the same prompts repeatedly.

\subsection{Experimental Setup}
Our initial step involved training a multi-abnormality classification model on all our available training data, comprising 20,000 unique patient profiles with 18 different abnormality labels, using real chest CT volumes. This baseline achieved a mean average precision (AP) of $0.254$ and a mean area under the receiver operating characteristic curve (AUROC) of $0.631$. To illustrate GenerateCT's effectiveness in scenarios with available real patient data, we augmented the training dataset by creating an equal number of synthetic volumes with GenerateCT, effectively doubling it. Furthermore, to demonstrate GenerateCT's efficacy in situations lacking real patient data and its capacity to generate large numbers of synthetic volumes, we produced 100,000 CT volumes, fivefold the number in our original dataset, through the repetitive use of the same prompts and trained the classifier solely on this synthetic data. Our experiment utilized the CT-Net model \cite{draelos2021machine}, with its default parameters for classifying 18 distinct abnormalities. The Stochastic Gradient Descent optimizer \cite{ruder2016overview} was employed with a learning rate of 0.001 and a weight decay of 0.0000001. All training sessions spanned 15 epochs with a batch size of 12, conducted on three A6000 48G GPUs. For consistency, all volumes were resized to $420\times420\times201$, and HU values were calibrated to a range of $[-1000, +200]$, focusing on heart and lung abnormalities \cite{denotter2019hounsfield}.

\subsection{Experimental Results}
\cref{table_aug_sup} details the model's performance in various training scenarios, highlighting AUROC and AP metrics across 18 abnormalities. We observed an $11\%$ improvement in mean AP and a $6\%$ increase in mean AUROC when training on both real and an equal number of synthetic volumes, compared to using only real data. Expanding the synthetic dataset to 100,000 volumes and training exclusively on this data resulted in an $8\%$ rise in mean AP and a $4\%$ increase in mean AUROC compared to the model trained on all the real data available to us. Validation was performed on the same real-patient dataset across all training scenarios. 

These results underscore GenerateCT's effectiveness in data augmentation; significantly enhancing performance by only doubling the dataset size proves beneficial for researchers with access to real-world data. Moreover, training with a larger, entirely synthetic dataset produced superior results over the real-data-only model, underscoring GenerateCT's role in ensuring data privacy. This approach facilitates the training and sharing of models like ours, allowing the generation of synthetic data using text prompts, thus enhancing performance while avoiding privacy or data-sharing concerns. Furthermore, the consistent improvement in performance metrics, even with repetitive use of the same prompts, illustrates GenerateCT's ability to produce varied data from identical inputs.

In conclusion, the results in \cref{table_aug_sup} establish GenerateCT as a valuable asset in data augmentation. Our experimental findings underscore GenerateCT's capability to generate detailed and realistic 3D chest CT volumes that accurately align with diverse text prompts. These outcomes mark a significant advancement in 3D medical imaging, suggesting that GenerateCT can be a powerful tool for enhancing diagnostic and treatment planning processes. Moreover, the potential of GenerateCT to simulate realistic, high-resolution medical images based on textual descriptions opens new avenues for future applications in healthcare.

\section{Utilizing GenerateCT in a Zero-Shot Setting}
\label{Zero}

In this section, we detail the application of GenerateCT in a zero-shot scenario, evaluating the model's ability to generalize to external datasets and perform with unseen prompts. We selected RadChestCT~\cite{draelos2021machine} as the external dataset, which comprises 3,630 3D chest CT volumes featuring 83 different abnormalities and a mean abnormality label frequency of $0.129$.

\subsection{Experimental Setup}
Initially, we established a baseline by training the classifier on RadChestCT, which included 2,286 3D CT volumes for training and 1,344 for validation. Each volume was associated with labels for 83 unique abnormalities. Subsequently, we generated a new dataset matching the volume count and abnormality distribution of RadChestCT's training set, resulting in 2,286 synthetic 3D CT volumes. The generation process employed structured medical language text prompts, \textbf{\textit{\{age\}~years~old~\{sex\}:~\{impression\}}}, where \{impression\} denoted the specific abnormalities. These text prompts were novel, not included in the original training data for GenerateCT, and featured a unique distribution of abnormalities. Due to the absence of age and sex parameters in RadChestCT, these were assigned randomly. The classifier underwent training using both the synthetic dataset and a combination of synthetic and real data. To ensure consistency, we applied the same preprocessing and model parameters as described in \cref{Augmentation}.

\subsection{Experimental Results}
\cref{table_zero_sup} presents the scores for each training scenario across all 83 abnormalities, noting comparable results between models trained on synthetic and real data: a mean AP of $0.146$ and AUROC of $0.536$ for synthetic, against $0.177$ AP and $0.613$ AUROC for real data. This similarity is significant, given that both scenarios used the same real patient dataset for validation, originating from a different institutional setup than that used for GenerateCT training. Training jointly with synthetic and real patient data showed a modest increase in both mean AUROC ($0.623$) and mean AP ($0.190$), underscoring the value of synthetic data in model training. The results in \cref{table_zero_sup} establish GenerateCT as a valuable tool for data generation from unseen prompts. Our experimental results highlight GenerateCT's ability to create detailed and realistic 3D chest CT volumes that correspond accurately to diverse text prompts not used during training. This demonstrates the extension of GenerateCT's key benefits, mentioned in \cref{Augmentation}, to external datasets and its potential for clinical application with unseen prompts.

\definecolor{mycolor}{rgb}{0.90, 0.90, 0.90} 

\begin{sidewaystable}
  \centering
  \caption{Performance metrics for all abnormalities with incremental data augmentation using GenerateCT. This highlights its significant clinical utility, especially in data augmentation, and its applicability in scenarios where real patient data sharing is challenging. It facilitates sharing trained models rather than private patient data, especially since the synthetic-only model outperforms the real data-only model.}
\vspace{0.0cm}
  \setlength{\tabcolsep}{2.3pt} 
  \renewcommand{\arraystretch}{1.2} 
  \begin{tabular}{
    @{}
    >{\raggedright\arraybackslash}p{4.0cm} 
    >{\columncolor{mycolor}}S[table-format=1.3]
    S[table-format=1.3]
    >{\columncolor{mycolor}}S[table-format=1.3]
    S[table-format=1.3]
    >{\columncolor{mycolor}}S[table-format=1.3]
    S[table-format=1.3]
    >{\columncolor{mycolor}}S[table-format=1.3]
    S[table-format=1.3]
    >{\columncolor{mycolor}}S[table-format=1.3]
    S[table-format=1.3]
    S[table-format=1.3]
    @{}
  }
    \toprule
    & \multicolumn{2}{c}{\cellcolor{white}\textbf{20k Real}} & \multicolumn{2}{c}{\cellcolor{white}\textbf{20k Real+20k Synth}} & \multicolumn{2}{c}{\cellcolor{white}\textbf{20k Synthetic}} & \multicolumn{2}{c}{\cellcolor{white}\textbf{40k Synthetic}} & \multicolumn{2}{c}{\cellcolor{white}\textbf{100k Synthetic}} & \cellcolor{white}\textbf{} \\
    \cmidrule(lr){2-3} \cmidrule(lr){4-5} \cmidrule(lr){6-7} \cmidrule(lr){8-9} \cmidrule(lr){10-11} 
    \textbf{Abnormality} & {\textbf{AUROC}} & {\textbf{AP}} & {\textbf{AUROC}} & {\textbf{AP}} & {\textbf{AUROC}} & {\textbf{AP}} & {\textbf{AUROC}} & {\textbf{AP}} & {\textbf{AUROC}} & {\textbf{AP}} & {\textbf{Test Set}} \\
    \midrule
    Medical material & 0.650 & 0.143 & 0.702 & 0.156 & 0.594 & 0.109 & 0.623 & 0.141 & 0.656 & 0.149 & 0.082 \\
    Arterial wall calcification & 0.648 & 0.434 & 0.605 & 0.452 & 0.714 & 0.445 & 0.705 & 0.415 & 0.715 & 0.405 & 0.292 \\
    Cardiomegaly & 0.804 & 0.310 & 0.745 & 0.352 & 0.590 & 0.142 & 0.593 & 0.154 & 0.599 & 0.184 & 0.102 \\
    Pericardial effusion & 0.667 & 0.044 & 0.685 & 0.056 & 0.642 & 0.079 & 0.690 & 0.094 & 0.662 & 0.152 & 0.026 \\
    C. artery wall calcification & 0.649 & 0.384 & 0.691 & 0.452 & 0.794 & 0.428 & 0.790 & 0.493 & 0.825 & 0.493 & 0.246 \\
    Hiatal hernia & 0.544 & 0.159 & 0.542 & 0.152 & 0.638 & 0.298 & 0.652 & 0.325 & 0.685 & 0.345 & 0.140 \\
    Lymphadenopathy & 0.616 & 0.345 & 0.679 & 0.399 & 0.591 & 0.301 & 0.612 & 0.351 & 0.642 & 0.345 & 0.245 \\
    Emphysema & 0.522 & 0.202 & 0.621 & 0.254 & 0.512 & 0.235 & 0.542 & 0.254 & 0.582 & 0.288 & 0.193 \\
    Atelectasis & 0.609 & 0.314 & 0.585 & 0.352 & 0.595 & 0.284 & 0.550 & 0.276 & 0.625 & 0.297 & 0.231 \\
    Lung nodule & 0.560 & 0.483 & 0.621 & 0.456 & 0.523 & 0.415 & 0.562 & 0.420 & 0.685 & 0.452 & 0.449 \\
    Lung opacity & 0.603 & 0.477 & 0.785 & 0.490 & 0.549 & 0.485 & 0.542 & 0.506 & 0.598 & 0.545 & 0.395 \\
    Pulmonary fibrotic sequela & 0.531 & 0.256 & 0.638 & 0.258 & 0.558 & 0.241 & 0.592 & 0.240 & 0.624 & 0.245 & 0.241 \\
    Pleural effusion & 0.777 & 0.323 & 0.815 & 0.365 & 0.632 & 0.198 & 0.678 & 0.205 & 0.725 & 0.286 & 0.125 \\
    Mosaic attenuation pattern & 0.739 & 0.152 & 0.712 & 0.195 & 0.594 & 0.097 & 0.612 & 0.087 & 0.661 & 0.125 & 0.056 \\
    Peribronchial thickening & 0.513 & 0.073 & 0.503 & 0.152 & 0.551 & 0.084 & 0.580 & 0.099 & 0.604 & 0.158 & 0.069 \\
    Consolidation & 0.655 & 0.235 & 0.693 & 0.264 & 0.592 & 0.154 & 0.599 & 0.168 & 0.6355 & 0.185 & 0.146 \\
    Bronchiectasis & 0.573 & 0.111 & 0.658 & 0.132 & 0.535 & 0.098 & 0.582 & 0.095 & 0.598 & 0.098 & 0.093 \\
    Interlobular septal thickening & 0.699 & 0.132 & 0.768 & 0.141 & 0.614 & 0.121 & 0.645 & 0.124 & 0.691 & 0.185 & 0.070 \\
    \midrule
    \textbf{Mean} & \textit{0.631} & \textit{0.254} & \textit{0.669} & \textit{0.282} & \textit{0.601} & \textit{0.234} & \textit{0.619} & \textit{0.247} & \textit{0.656} & \textit{0.274} & \textit{0.179}\\
    \bottomrule
  \end{tabular}
  \label{table_aug_sup}
\end{sidewaystable}

\begin{table}[ht]
  \centering
  \tiny
  \caption{Performance metrics for different abnormalities across various training datasets, highlighting GenerateCT's generation capability based on unseen prompts.}
\vspace{-0.3cm}
  \setlength{\tabcolsep}{4.5pt} 
  \renewcommand{\arraystretch}{0.95} 
  \begin{tabular}{
    @{}
    >{\raggedright\arraybackslash}p{3.0cm} 
    >{\columncolor{mycolor}}S[table-format=1.3]
    S[table-format=1.3]
    >{\columncolor{mycolor}}S[table-format=1.3]
    S[table-format=1.3]
    >{\columncolor{mycolor}}S[table-format=1.3]
    S[table-format=1.3]
    S[table-format=1.3]
    @{}
  }
    \toprule
    & \multicolumn{2}{c}{\cellcolor{white}\textbf{Real Data}} & \multicolumn{2}{c}{\cellcolor{white}\textbf{Synthetic Data}} & \multicolumn{2}{c}{\cellcolor{white}\textbf{Composite Data}} & \cellcolor{white}\textbf{} \\
    \cmidrule(lr){2-3} \cmidrule(lr){4-5} \cmidrule(lr){6-7}
    \textbf{Abnormality} & {\textbf{AUROC}} & {\textbf{AP}} & {\textbf{AUROC}} & {\textbf{AP}} & {\textbf{AUROC}} & {\textbf{AP}} & {\textbf{Test Set}} \\
    \midrule
    Air trapping               & 0.561 & 0.044 & 0.621 & 0.050 & 0.633 & 0.051 & 0.031 \\
    Airspace disease           & 0.605 & 0.258 & 0.571 & 0.210 & 0.607 & 0.233 & 0.171 \\
    Aneurysm                   & 0.577 & 0.015 & 0.493 & 0.012 & 0.587 & 0.020 & 0.011 \\
    Arthritis                  & 0.515 & 0.284 & 0.510 & 0.298 & 0.505 & 0.282 & 0.279 \\
    Aspiration                 & 0.616 & 0.091 & 0.518 & 0.051 & 0.624 & 0.092 & 0.049 \\
    Atelectasis                & 0.575 & 0.349 & 0.579 & 0.356 & 0.596 & 0.408 & 0.290 \\
    Atherosclerosis            & 0.550 & 0.314 & 0.473 & 0.281 & 0.525 & 0.297 & 0.294 \\
    Bandlike or linear         & 0.461 & 0.156 & 0.511 & 0.191 & 0.483 & 0.166 & 0.177 \\
    Breast implant             & 0.387 & 0.016 & 0.325 & 0.012 & 0.550 & 0.066 & 0.017 \\
    Breast surgery             & 0.499 & 0.030 & 0.504 & 0.026 & 0.484 & 0.037 & 0.023 \\
    Bronchial thickening       & 0.556 & 0.080 & 0.474 & 0.074 & 0.566 & 0.086 & 0.070 \\
    Bronchiectasis             & 0.704 & 0.313 & 0.543 & 0.179 & 0.666 & 0.234 & 0.154  \\
    Bronchiolectasis           & 0.739 & 0.068 & 0.475 & 0.021 & 0.683 & 0.044 & 0.021 \\
    Bronchiolitis              & 0.443 & 0.024 & 0.492 & 0.025 & 0.509 & 0.026 & 0.025 \\
    Bronchitis                 & 0.533 & 0.010 & 0.567 & 0.023 & 0.570 & 0.011 & 0.008 \\
    CABG                       & 0.754 & 0.118 & 0.504 & 0.068 & 0.764 & 0.115 & 0.041  \\
    Calcification              & 0.426 & 0.669 & 0.501 & 0.727 & 0.428 & 0.676 & 0.721  \\
    Cancer                     & 0.593 & 0.614 & 0.523 & 0.575 & 0.618 & 0.636 & 0.563  \\
    Cardiomegaly               & 0.752 & 0.238 & 0.622 & 0.142 & 0.798 & 0.314 & 0.094  \\
    Catheter or port           & 0.660 & 0.218 & 0.591 & 0.120 & 0.681 & 0.266 & 0.084 \\
    Cavitation                 & 0.604 & 0.056 & 0.493 & 0.058 & 0.589 & 0.056 & 0.040 \\
    Chest tube                 & 0.864 & 0.123 & 0.640 & 0.037 & 0.881 & 0.173 & 0.018  \\
    Clip                       & 0.488 & 0.098 & 0.532 & 0.117 & 0.491 & 0.106 & 0.092 \\
    Congestion                 & 0.885 & 0.042 & 0.701 & 0.015 & 0.951 & 0.266 & 0.005 \\
    Consolidation              & 0.690 & 0.286 & 0.565 & 0.193 & 0.680 & 0.256 & 0.139 \\
    Coronary artery disease    & 0.567 & 0.608 & 0.500 & 0.568 & 0.582 & 0.607 & 0.566 \\
    Cyst                       & 0.497 & 0.169 & 0.469 & 0.156 & 0.488 & 0.162 & 0.167 \\
    Debris                     & 0.697 & 0.081 & 0.572 & 0.048 & 0.697 & 0.111 & 0.038 \\
    Deformity                  & 0.580 & 0.062 & 0.475 & 0.051 & 0.551 & 0.057 & 0.052 \\
    Density                    & 0.536 & 0.106 & 0.499 & 0.095 & 0.538 & 0.116 & 0.092  \\
    Dilation or ectasia        & 0.571 & 0.063 & 0.458 & 0.051 & 0.589 & 0.066 & 0.046 \\
    Distention                 & 0.592 & 0.020 & 0.653 & 0.056 & 0.641 & 0.019 & 0.011 \\
    Emphysema                  & 0.623 & 0.329 & 0.421 & 0.230 & 0.614 & 0.352 & 0.275 \\
    Fibrosis                   & 0.792 & 0.332 & 0.574 & 0.152 & 0.775 & 0.259 & 0.118 \\
    Fracture                   & 0.601 & 0.094 & 0.536 & 0.075 & 0.588 & 0.097 & 0.070 \\
    GI tube                    & 0.900 & 0.192 & 0.710 & 0.067 & 0.910 & 0.269 & 0.018 \\
    Granuloma                  & 0.448 & 0.071 & 0.411 & 0.066 & 0.450 & 0.071 & 0.080 \\
    Groundglass                & 0.594 & 0.415 & 0.524 & 0.341 & 0.589 & 0.422 & 0.325 \\
    Hardware                   & 0.447 & 0.022 & 0.513 & 0.028 & 0.416 & 0.021 & 0.026 \\
    Heart failure              & 0.878 & 0.056 & 0.585 & 0.013 & 0.951 & 0.199 & 0.009 \\
    Heart valve replacement    & 0.745 & 0.043 & 0.721 & 0.059 & 0.858 & 0.165 & 0.014 \\
    Hemothorax                 & 0.889 & 0.125 & 0.721 & 0.011 & 0.833 & 0.032 & 0.005 \\
    Hernia                     & 0.523 & 0.120 & 0.488 & 0.126 & 0.548 & 0.126 & 0.115  \\
    Honeycombing               & 0.903 & 0.258 & 0.566 & 0.045 & 0.846 & 0.105 & 0.032 \\
    Infection                  & 0.539 & 0.355 & 0.448 & 0.301 & 0.538 & 0.354 & 0.317 \\
    Infiltrate                 & 0.413 & 0.015 & 0.438 & 0.021 & 0.352 & 0.014 & 0.018 \\
    Inflammation               & 0.529 & 0.087 & 0.429 & 0.076 & 0.521 & 0.086 & 0.082 \\
    Interstitial lung disease  & 0.739 & 0.362 & 0.565 & 0.196 & 0.742 & 0.304 & 0.152 \\
    Lesion                     & 0.467 & 0.234 & 0.487 & 0.246 & 0.482 & 0.235 & 0.251 \\
    Lucency                    & 0.574 & 0.028 & 0.567 & 0.028 & 0.556 & 0.041 & 0.018 \\
    Lung resection             & 0.519 & 0.222 & 0.516 & 0.236 & 0.545 & 0.242 & 0.229 \\
    Lymphadenopathy            & 0.682 & 0.260 & 0.580 & 0.191 & 0.686 & 0.272 & 0.151 \\
    Mass                       & 0.498 & 0.123 & 0.541 & 0.149 & 0.505 & 0.128 & 0.128  \\
    Mucous plugging            & 0.519 & 0.028 & 0.413 & 0.027 & 0.480 & 0.027 & 0.028 \\
    Nodule                     & 0.649 & 0.858 & 0.600 & 0.855 & 0.682 & 0.873 & 0.800 \\
    Nodule \textgreater 1cm    & 0.515 & 0.136 & 0.544 & 0.158 & 0.499 & 0.121 & 0.128 \\
    Opacity                    & 0.369 & 0.456 & 0.539 & 0.571 & 0.634 & 0.667 & 0.543 \\
    Pacemaker/defibrillator    & 0.778 & 0.128 & 0.563 & 0.079 & 0.857 & 0.261 & 0.049 \\
    Pericardial effusion       & 0.626 & 0.207 & 0.544 & 0.167 & 0.629 & 0.236 & 0.143  \\
    Pericardial thickening     & 0.501 & 0.024 & 0.551 & 0.076 & 0.538 & 0.026 & 0.025 \\
    Plaque                     & 0.608 & 0.034 & 0.408 & 0.023 & 0.566 & 0.031 & 0.024 \\
    Pleural effusion           & 0.770 & 0.424 & 0.656 & 0.308 & 0.792 & 0.507 & 0.199 \\
    Pleural thickening         & 0.583 & 0.120 & 0.573 & 0.125 & 0.549 & 0.118 & 0.100 \\
    Pneumonia                  & 0.629 & 0.079 & 0.569 & 0.067 & 0.664 & 0.096 & 0.050 \\
    Pneumonitis                & 0.677 & 0.070 & 0.578 & 0.034 & 0.689 & 0.052 & 0.027 \\
    Pneumothorax               & 0.780 & 0.196 & 0.576 & 0.030 & 0.815 & 0.193 & 0.024 \\
    Postsurgical               & 0.554 & 0.525 & 0.517 & 0.503 & 0.537 & 0.521 & 0.485 \\
    Pulmonary edema            & 0.816 & 0.144 & 0.638 & 0.081 & 0.852 & 0.217 & 0.034 \\
    Reticulation               & 0.747 & 0.211 & 0.559 & 0.121 & 0.710 & 0.165 & 0.090 \\
    Scarring                   & 0.448 & 0.193 & 0.462 & 0.219 & 0.531 & 0.247 & 0.227 \\
    Scattered calcifications   & 0.519 & 0.187 & 0.506 & 0.190 & 0.491 & 0.187 & 0.183 \\
    Scattered nodules          & 0.497 & 0.216 & 0.463 & 0.211 & 0.494 & 0.225 & 0.223  \\
    Secretion                  & 0.587 & 0.019 & 0.530 & 0.019 & 0.599 & 0.021 & 0.014 \\
    Septal thickening          & 0.793 & 0.176 & 0.612 & 0.105 & 0.794 & 0.195 & 0.060 \\
    Soft tissue                & 0.475 & 0.166 & 0.558 & 0.206 & 0.466 & 0.160 & 0.171 \\
    Staple                     & 0.501 & 0.032 & 0.536 & 0.040 & 0.462 & 0.033 & 0.031  \\
    Stent                      & 0.580 & 0.040 & 0.550 & 0.064 & 0.554 & 0.037 & 0.032 \\
    Sternotomy                 & 0.743 & 0.186 & 0.536 & 0.086 & 0.779 & 0.241 & 0.068 \\
    Suture                     & 0.507 & 0.028 & 0.534 & 0.022 & 0.466 & 0.022 & 0.020 \\
    Tracheal tube              & 0.937 & 0.234 & 0.710 & 0.033 & 0.931 & 0.232 & 0.013 \\
    Transplant                 & 0.701 & 0.174 & 0.574 & 0.099 & 0.713 & 0.178 & 0.074 \\
    Tree in bud                & 0.573 & 0.064 & 0.399 & 0.020 & 0.591 & 0.035 & 0.023 \\
    Tuberculosis               & 0.534 & 0.005 & 0.366 & 0.003 & 0.467 & 0.006 & 0.003 \\
    \midrule
    \textbf{Mean} & \textit{0.613} & \textit{0.177} & \textit{0.536} & \textit{0.146} & \textit{0.623} & \textit{0.190} & \textit{0.129}\\
    \bottomrule
  \end{tabular}
  \label{table_zero_sup}
\end{table}

\end{document}